# Curriculum Adversarial Training

**Qi-Zhi Cai**[1*]  **Min Du**[2]  **Chang Liu**[3]  **Dawn Song**[3]

[1] Nanjing University   [2] University of Utah   [3] UC Berkeley

## Abstract

Recently, deep learning has been applied to many security-sensitive applications, such as facial authentication. The existence of adversarial examples hinders such applications. The state-of-the-art result on defense shows that adversarial training can be applied to train a robust model on MNIST against adversarial examples; but it fails to achieve a high empirical worst-case accuracy on a more complex task, such as CIFAR-10 and SVHN. In our work, we propose *curriculum adversarial training* (CAT) to resolve this issue. The basic idea is to develop a curriculum of adversarial examples generated by attacks with a wide range of strengths. With two techniques to mitigate the *catastrophic forgetting* and the *generalization* issues, we demonstrate that CAT can improve the prior art's empirical worst-case accuracy by a large margin of 25% on CIFAR-10 and 35% on SVHN. At the same, the model's performance on non-adversarial inputs is comparable to the state-of-the-art models.

## 1 Introduction

Recently, deep learning has achieved the state-of-the-art performance on many tasks, such as images classification [He *et al.*, 2016], game playing [Silver *et al.*, 2016], and speech recognition [Hinton *et al.*, 2012], and has been applied to many security- and safety-sensitive applications, such as autonomous driving [Chen *et al.*, 2015] and facial authentication [Sun *et al.*, 2014]. However, the existence of *adversarial examples* severely hinders the application of deep learning to these problems [Szegedy *et al.*, 2013; Goodfellow *et al.*, 2015].

Since [Szegedy *et al.*, 2013], many efforts have been devoted to investigate different attacks leveraging adversarial examples [Papernot *et al.*, 2016a; Goodfellow *et al.*, 2015; Liu *et al.*, 2017; Carlini and Wagner, 2017b; Moosavi-Dezfooli *et al.*, 2017] and defenses [Papernot *et al.*, 2016b; Kurakin *et al.*, 2017; Madry *et al.*, 2018; Jacob Buckman, 2018; Xu *et al.*, 2018]. Although many defense strategies have been proposed, almost all of them are shown to be broken when the attacker can adapt the attack to take the defense into consideration [Carlini and Wagner, 2017b; 2017a; He *et al.*, 2017]. It has been several years that the community is unclear whether building a robust model is even possible.

A recent work from [Madry *et al.*, 2018] provides the first positive answer. The idea is straightforward: during training, the model is trained with not only standard training data, but also adversarial examples generated from an attack. Such an approach is referred to as *adversarial training*, and has been exploited by many existing works [Goodfellow *et al.*, 2015; Kurakin *et al.*, 2017; Li *et al.*, 2016]. [Madry *et al.*, 2018] provide two insights on why previous adversarial training approaches cannot train a robust model: (1) the model should have a sufficiently large capacity; and (2) strong attacks should be used to generate the adversarial examples during training. In doing so, [Madry *et al.*, 2018] show that, slightly imprecisely speaking, adversarial training can result in a robust model for MNIST [LeCun, 1998] with an *empricial worst-case test accuracy* of around 90%. That is, for around 90% of the test data, no known attacks can generate effective adversarial examples. Although this result is encouraging, the approach does not generalize well to a slightly more difficult task, such as CIFAR-10 [Krizhevsky, 2009]. In particular, [Madry *et al.*, 2018] can only achieve an empirical worst-case test accuracy of around 45% on CIFAR-10 and 40% on SVHN.

In this work, we follow the line of research on the adversarial training framework, and design a training technique, called *curriculum adversarial training* (CAT). Compared with [Madry *et al.*, 2018], CAT is both more effective and more efficient. We offer two novel insights on adversarial training: (1) although [Madry *et al.*, 2018] show that strong attacks should be used in an adversarial training framework, we show that weak attacks are also useful to mitigate a *catastrophic forgetting* issue; and (2) quantization can effectively help our CAT technique to achieve better *generalization*. Note that, quantization has been exploited in existing works as a defense [Xu *et al.*, 2018; Jacob Buckman, 2018]; however, [He *et al.*, 2017] and [Anish Athalye, 2018] show that quantization alone can be easily broken. Our work is the first to demonstrate that quantization can significantly improve the robustness of a model against adversarial examples using the CAT approach.

We evaluate our approach on CIFAR-10 and SVHN [Netzer *et al.*, 2011], and compare our approach against the state-of-the-art approach from [Madry *et al.*, 2018]. We observe that our approach can consistently improve the empirical worst-case accuracy: from 46.18% to 69.27% on CIFAR-10, and from 40.38% to 75.66% on SVHN. Also, on non-adversarial test data, the accuracy of models trained using our approach decreases the performance of the state-of-the-art models by at most 6%. Therefore, CAT has potentials to be deployed in practice to achieve a robust model.

**Related Work.** Since adversarial examples were first demonstrated [Szegedy *et al.*, 2013; Goodfellow *et al.*, 2015], it has been several years that researchers are forming an army race to propose attacks [Papernot *et al.*, 2016a; Liu *et al.*, 2017; Carlini and Wagner, 2017b; Moosavi-Dezfooli *et al.*, 2017] and defenses [Papernot *et al.*, 2016b; Kurakin *et al.*, 2017; Madry *et al.*, 2018; Jacob Buckman, 2018; Xu *et al.*, 2018]. To date, most defenses have been shown to be broken [Carlini and Wagner, 2017b; 2017a; He *et al.*, 2017].

---

[*] The work is done at UC Berkeley

Despite such a negative result, [Madry *et al.*, 2018] propose an adversarial training approach, and no one has shown any effective attacks to break the MNIST model trained using this approach so far. However, [Madry *et al.*, 2018] have not shown that the approach can be extended to handle a more complex task such as CIFAR-10 or SVHN. Our work follows this direction to improve the adversarial training framework on these harder tasks in terms of both efficiency and empirical worst-case accuracy against adversarial examples.

It is worth mentioning that, since [Madry *et al.*, 2018], researchers also consider an alternative line to construct a model that is provably secure by construction [Kolter and Wong, 2017; Aditi Raghunathan, 2018; Aman Sinha, 2018]. These approaches exploit the structure of the neural network, and thus heavily rely on the choice of the neural network architecture. Due to this limitation, these approaches are only shown to be provably robust on MNIST, and it is unclear how to apply them to more complex tasks such as CIFAR-10.

## 2 Adversarial examples and adversarial training

In this section, we first present the problem of finding adversarial examples, and then present the generic framework of adversarial training to train a robust model against adversarial examples.

### 2.1 Adversarial examples

Given a network $f_\theta(\cdot)$ and an input $x$, whose ground truth label is $y$, an adversarial example $x^\star$ is an input such that

$$f_\theta(x^\star) \neq y \wedge d(x, x^\star) \leq \epsilon \quad (1)$$

Here, $d(\cdot, \cdot)$ is a distance metric to quantify the *semantic distance* between the two inputs $x$ and $x^\star$ is small enough, so that they share the same ground truth.

As in most existing works, we focus on adversarial examples to the image domain. That is, the input space contains all images of the same size. Therefore, the two images $x$ and $x^\star$ can be represented as two vectors. There have been some distance metrics considered in the literature, such as $L_p$ norm of $x - x^\star$ for $p = 0$ [Carlini and Wagner, 2017b], $p = 1$ [Sharma and Chen, 2018], $p = 2$ [Moosavi-Dezfooli *et al.*, 2017; Carlini and Wagner, 2017b], and $p = \infty$ [Madry *et al.*, 2018; Goodfellow *et al.*, 2015]. In this work, we follow [Madry *et al.*, 2018] to consider $d(x, x') = L_\infty(x - x^\star)$. However, we emphasize that our approach can be easily adapted to other distance metrics.

### 2.2 Adversarial example attacks

Finding an adversarial example is equivalent to solving the satisfiability problem of (1). However, directly solving it is hard. Most existing adversarial example attacks convert (1) into the following optimization problem:

$$\operatorname*{argmax}_{x^\star} \mathcal{L}(f_\theta(x^\star), y) \quad (2)$$

$$\text{s.t. } d(x, x^\star) \leq \epsilon \quad (3)$$

Here, $\mathcal{L}$ is a loss function between a prediction $f_\theta(x^\star)$ and the ground truth $y$. Intuitively, once the loss function is maximized, it is more likely that $f_\theta(x^\star) \neq y$.

Different approaches have been proposed to solve the above optimization problem. For the $L_\infty$ norm distance metric, we explain two state-of-the-art approaches below, *Iterative optimization attack* [Carlini and Wagner, 2017b] and *Projected Gradient Descent* (PGD) [Madry *et al.*, 2018]. We will also explain the strength of attacks, and the idea of black-box attacks, which are effective at bypassing many defenses.

**Iterative optimization attack.** The Carlini and Wagner's $L_\infty$ attack optimizes the following term

$$\operatorname*{argmin}_{x^\star} -\xi \mathcal{L}(f_\theta(x^\star), y) + \sum_i \text{ReLU}((x^\star - x)_i - \tau)$$

where $\xi$ is a hyper-parameter to be empirically determined, and $\tau$ is initialized with $\epsilon$, and decreased by a factor of $0.9\times$ when the second term becomes zero.

Intuitively, minimizing the first term is equivalent to maximizing (2), while when second term is minimized, i.e., to be 0, then the constraint (3) is satisfied. $\xi$ is a hyper-parameter to balance the two terms in the objective. The iterative optimization attack is generally regarded as the state-of-the-art attack.

**Projected Gradient Descent (PGD).** A PGD($k$) algorithm is parameterized with $k$, indicating the number of iterations. The basic idea is to optimize (2) using an iterative gradient descent algorithm:

$$x^{i+1} \leftarrow \Omega\big(x^i + \frac{\epsilon}{k} \mathbf{sgn}\left(\nabla_x \mathcal{L}(f_\theta(x^i), y)\right)$$

where $x^0$ is set with $x$, and $\mathbf{sgn}(\cdot)$ converts each dimension of a vector to its sign (i.e., $\{-1, 0, +1\}$). $\Omega(\cdot)$ is a projection function to ensure that each dimension of the input is projected to be in a valid range (i.e., $[0, 1]$): $\Omega(v)_i = \text{clip}(v_i, 0, 1)$. Since $||\mathbf{sgn}(\cdot)||_\infty \leq 1$ holds for any input, it is easy to show that $||x^i - x^0||_\infty \leq i\epsilon/k$. Therefore, the final output $x^\star = x^k$ satisfies $||x^\star - x||_\infty \leq \epsilon$, which is constraint (3). It is worth noting that an earlier approach called *Fast Gradient Sign* (FGS) [Goodfellow *et al.*, 2015] is a special instance of PGD(1).

**Attack strength.** Now we take a detour to introduce the concepts of *weak attacks* versus *strong attacks*. Intuitively, weak attacks refer to those attacks that are easy to defend against, while strong attacks are hard. For example, FGS [Goodfellow *et al.*, 2015] is generally regarded as a weak attack, and many defenses [Papernot *et al.*, 2016b; Kurakin *et al.*, 2017] have been demonstrated effective against FGS attacks. On the other hand, the iterative optimization attack is considered *strong*, and it is shown to be effective at breaking many defenses against FGS.

Mathematically, all attacks can be viewed as an approximation to the optimum of Objective (2) under Constraint (3). Thus, the *strength* of an attack can be measured by how close it can approximate the true optimum. From this point of view, PGD($k$) can be viewed as a class of attacks whose strength is parameterized by $k$. That is, the larger the value of $k$ is, a finer-grained approximation the result is, and thus a stronger attack it is.

Our curriculum adversarial training framework relies on the existence of an attack class whose strength can be parameterized. Clearly, by parameterizing the number of iterations used to generate an adversarial example, we can easily create different attack classes for other distance metrics as well. Therefore, although we focus on $L_\infty$ norm distance, our approach can be adapted to other distance metrics as well.

**Black-box attacks.** All above discussed attacks assume that the parameters of the model $\theta$ is known, so that a derivation of $\mathcal{L}$ can be computed. We refer to such an attack as *white-box* attacks. When an attack does not rely on the knowledge of $\theta$, we call them *black-box* attacks. There have been several approaches to show black-box attacks are possible due to transferability [Papernot *et al.*, 2016a; Liu *et al.*, 2017]. The basic idea is that, the attacker can train a substitute model $f'_{\theta'}$ on the same task and generates adversarial examples against $f'_{\theta'}$; then these adversarial examples will also be misclassified by the target model $f_\theta$ even though the model $f'_{\theta'}$ may be very different from the target $f_\theta$. Such a black-box attack can bypass many defenses that rely on making the model non-differentiable to prevent gradient-based attacks.

### 2.3 Adversarial training

A natural idea for building a robust model is to minimize the upper bound of (2) for all inputs. That is, finding the saddle point of the following objective [Madry *et al.*, 2018]:

$$\operatorname*{argmin}_{\theta} \max_{(x,y) \in \mathcal{D}} \max_{d(x,x^\star) \leq \epsilon} \mathcal{L}(f_\theta(x^\star), y) \quad (4)$$

where $\mathcal{D}$ is a dataset (e.g., a training set or a test set). The basic idea for adversarial retraining is to solve (4) using an alternative optimization approach. That is, the algorithm iteratively does the following two steps:

1. Fixing all $x^\star$, optimize $\theta$ for the outer minimization problem; and
2. Fixing $\theta$, finding the worst-case adversarial examples $x^\star$ for the inner maximization problem.

This procedure is called an *adversarial training algorithm*, and summarized in Algorithm 1.

Albeit its conceptual simplicity, earlier attempts instantiating the algorithm using FGS as $\mathcal{G}$ do not result in a robust model [Kurakin *et al.*, 2017]. Only until recently, [Madry *et al.*, 2018] shows that adversarial training can be used to obtain a robust MNIST model. That is, the worst-case accuracy on MNIST test data can be no lower than 88% even though the attacker knows all parameters of the model. The two key observations from Madry et al. are (1) strong attacks should be used in the second step to obtain a robust model; and (2) the model should have a sufficiently large capacity (e.g., the number of parameters). In particular, [Madry *et al.*, 2018] instantiates $\mathcal{G}$ with PGD(40) to achieve a robust MNIST model.

Although their results are encouraging, when the algorithm is extended to handle more complex tasks, such as CIFAR-10 or SVNH, the vanilla adversarial training algorithm from [Madry *et al.*, 2018] cannot achieve an empirical worst-case accuracy higher than 46% on CIFAR-10 and 40% on SVHN.

---

**Algorithm 1** Adversarial Training (AT($\mathcal{D}, N, \eta, \mathcal{G}$))

**Input:** Training data $\mathcal{D}$; Total iterations $N$; Learning rate $\eta$
**Input:** An attack $\mathcal{G}$
**Output:** $\theta$
1: Randomly initialize network $\theta$
2: **for** $i \leftarrow 0$ to $N$ **do**
3:     Sample a batch $(x_i, y_i) \sim \mathcal{D}$
4:     Generate adversarial examples $x_i^\star \leftarrow \mathcal{G}(x_i, y_i)$
5:     $\theta \leftarrow \theta - \eta \sum_i \nabla_\theta \mathcal{L}(f_\theta(x_i^\star), y_i)$
       // This is standard SGD; it can be replaced by
       other training algorithms such as Adam
6: **end for**

---

## 3 Curriculum adversarial training

In this section, we present our optimized adversarial training approach, called *curriculum adversarial training* (CAT). In the following, we first present the basic idea of curriculum adversarial training (Section 3.1), and then present two optimizations, including batch mixing (Section 3.2) and quantization (Section 3.3), for better performance.

### 3.1 Basic curriculum adversarial training

The problem with Madry et al.'s adversarial training algorithm is that it starts training the model from a strong attack generated by PGD($k$), (i.e., $k = 40$ for MNIST and $k = 7$ for CIFAR-10). In each training iteration, since it always finds the "worst-case" input using a strong attack, adversarial training seems to overfit to the attack in use and does not generalize to test data. As a result, Madry et al.'s approach is hard to achieve a high test accuracy on test data.

Our basic idea to mitigate these issues is to design a training curriculum. That is, the model is first trained with a *weak* attack; once the model can achieve a high accuracy against attacks at some strength, the attack strength is increased. This process continues until the attack strength reaches an upper bound specified by the defender.

The entire procedure is described in Algorithm 2. In this algorithm, $\mathcal{A}(k)$ denotes an attack class parameterized with $k$ indicating the attack strength. In our work, we instantiate $\mathcal{A}$ with PGD. We use AT to denote the standard adversarial training algorithm (Algorithm 1). The parameter $n$ is set to be $|\mathcal{D}|/\text{batch} - \text{size}$, so that each adversarial training call on line 4 is doing one epoch of adversarial training over $\mathcal{D}$.

Each iteration of the main loop (line 3-6) is called a *lesson* $l$ in the curriculum. In the first lesson $l = 0$, the model is trained with $\mathcal{A}(0)$, which indicates that the training data is directly fed into the training algorithm without generating any adversarial examples. After each epoch, we evaluate the $\tilde{l}$-accuracy on $\mathcal{V}$ and if it is not increased for the last 10 epoches, we increase the attack strength $l$. Here, we use $\tilde{k}$-accuracy to denote the percentage of examples in the validation set that no attacks in $\mathcal{A}(i)$ for $i = 0, ..., l$ can generate successful adversarial examples. Formally, we have

$$\tilde{k}\text{-accuracy}(\mathcal{V}, \theta)$$
$$= \frac{|\{(x,y) \in \mathcal{V} | \forall i \in \{0, ..., k\}. f_\theta(\mathcal{A}(i)(x)) = y\}|}{|\mathcal{V}|} \quad (5)$$

**Algorithm 2** Curriculum Adversarial Training (Basic)
---
**Input:** Training data $\mathcal{D}$; Validation data $\mathcal{V}$; Epoch iterations $n$; Learning rate $\eta$; Maximal attack strength $K$;
**Input:** An class of attacks, denoted as $\mathcal{A}(k)$ whose strength is parameterized by $k$.
**Output:** $\theta$
1: Randomly initialize network $\theta$
2: **for** $l \leftarrow 0 \ to \ K$ **do**
3:    **repeat**
4:       $\theta \leftarrow \text{AT}(\mathcal{D}, n, \eta, \mathcal{A}(l))$
5:       // One epoch of adversarial training using $\mathcal{A}(l)$
6:    **until** $\tilde{l}$-accuracy on $\mathcal{V}$ not increased for 10 epochs
7: **end for**
---

When the model's validation $\tilde{k}$-accuracy has not been updated in the last 10 epochs, we reset the parameters of the model to be the best ones (i.e., 10 epochs ago) on the validation set before increasing $k$.

The basic idea is that, while training adversarially with attacks $\mathcal{A}(k)$ at a certain strength $k$, we evaluate all attacks no stronger than $\mathcal{A}(k)$ on the validation set to moniter whether the model is overfitting. Once the model achieves the best validation performance using the current attack strength, we increase the attack strength to supervise the model with more challenging tasks. In doing so, we adversarially train the model with a designed curriculum from weaker attacks to stronger attacks.

### 3.2 Batch mixing

Although the basic curriculum training can achieve a significantly reduction on the training time, it does not increase the robustness. We observe that one issue is *forgetting*: when the model is trained with a stronger attack (i.e., for a larger $l$), it will *forget* the adversarial examples generated for a smaller $l$. As a result, the basic curriculum training cannot achieve a high empirical worst-case accuracy.

To mitigate this problem, when training with a strong attack, we also need to make the model remember the adversarial examples generated by weaker attacks. That is, to form a batch, instead of generating all adversarial examples using PGD($l$), we generate some adversarial examples using PGD($i$) for each $i \in \{0, 1, ..., l\}$, and combine them to form a batch. We refer to such a technique as *batch mixing*.

As a result, the loss function to compute its gradient (line 5 Algorithm 1) is

$$\sum_{i=0}^{k} \alpha_i \sum_{(x,y) \sim \mathcal{D}} \mathcal{L}(f_\theta(\mathcal{A}(i)(x)), y)$$

where $\alpha_i$ are the hyper-parameters such that $\forall i. \alpha_i \in [0, 1]$ and $\sum_{i=0}^{l} \alpha_i = 1$. In practice, we observe that setting $\alpha_i = \frac{1}{l+1}$ and generating the same amount of adversarial examples for each attack strength are effective enough. With batch mixing, we observe that the model trained with a strong attack can also remember most adversarial examples for weaker attacks, and thus yield a better overall accuracy.

### 3.3 Quantization

Another problem with curriculum adversarial training is *attack generalization*. That is, the model trained with CAT may not defend against attacks that are stronger than the strongest attack used during training. We mitigate this problem by employing *quantization*. Intuitively, each input $x$ is a vector of real values from $[0, 1]$. However, we can restrict each input dimension to be a $b$-bit integer rather than a real value. This technique is referred to as *quantization*.

In particular, quantization is an inference time technique to reduce the space of adversarial examples. Given any test input, the model will first convert each pixel value into a $b$-bit integer, before feeding it into the neural network. Without quantization, the difference of $x^\star - x$ can take any real value from $[-\epsilon, +\epsilon]^m$, where $m$ is the input dimension. This is an infinite space. In contrast, when the inputs are quantized to be $b$-bit integer vectors, $2^b x^\star$ can only take an integer value from $[\lceil 2^b(x-\epsilon) \rceil, \lceil 2^b(x+\epsilon) \rceil]^m$, which is a finite space. We refer to this space as the *adversarial example space*. Therefore, the smaller $b$ is, the smaller the adversarial example space is but the less effective the model may be, since the input may contain too little information. Therefore, the choice of $b$ is a trade-off between resilience against adversarial examples, and the effectiveness of the model. In our evaluation, we observe that choosing $b = 4$ is the best choice.

Quantiziation has been proposed by [Xu *et al.*, 2018] as one of the feature squeezing techniques. However, this technique alone is not shown to provide resilience against strong white-box attacks. When used in the adversarial training framework, in contrast, CAT can effectively train the model to remember as many instances in the adversarial example space as possible, with a reasonably weak attack. That is, although a stronger attack can better optimize the loss function, the adversarial examples that it generates are highly likely to coincide with those generated by a weaker attack, because the entire adversarial example space is small. In our evaluation, we observe that quantiziation can effectively improve the accuracy against stronger attacks than the model is observed during training.

## 4 Evaluation Setup

In this work, we evaluate our approach against the state-of-the-art approach [Madry *et al.*, 2018] on CIFAR-10, and SVHN. As we mentioned in Section 2, we use $L_\infty$ to derive the distance between benign data and adversarial examples. We set the hyper-parameters for different data sets to be the same as used in the literature:

For both datasets, we use two state-of-the-art image classification architectures: ResNet-50 [He *et al.*, 2016] and DenseNet-161 [Huang *et al.*, 2017]. Mini-batch size is set to be 200 for all our approaches.

|  | CIFAR-10 | | SVHN | |
| --- | --- | --- | --- | --- |
|  | ResNet-50 | DenseNet-161 | ResNet-50 | DenseNet-161 |
| Ours | 65.93% | **69.27%** | **75.66%** | 74.58% |
| [Madry *et al.*, 2018] | 45.39% | 46.18% | 19.59% | 40.38% |

Table 1: Empricial worst-case accuracy using our approach versus baselines on CIFAR-10 and SVHN. "Ours" indicate the CAT approach with all optimizations applied. The **best results** are in bold.

| Dataset | Bound | $K$ | used in |
| --- | --- | --- | --- |
| CIFAR-10 | 8/255 | 7 | [Madry *et al.*, 2018] |
| SVHN | 12/255 | 10 | [Jacob Buckman, 2018] |

**Empirical worst-case accuracy.** Although we have mentioned it several times, we formally define *empirical worst-case accuracy* now. In particular, for each test input $x$, we generate adversarial examples using different algorithms; if one of these adversarial examples can mislead the evaluated model, we mark input $x$ as *failed*. Then *empirical worst-case accuracy* is evaluated as the percentage of test cases that do not fail. Note that this approach does not guarantee to find real worst case adversarial example; however, it empirically approximates the worst case when a wide range of attacks are employed. In particular, we include the following attacks.

- The iterative optimization attack [Carlini and Wagner, 2017b];
- PGD($k$) for $k \in \{1, 2, 3, 4, 5, 6, 7, 8, 9, 10, 20, 50, 100\}$;
- Black-box attacks.

We also refer to empirical worst-case accuracy as *resilience* for short, and use the two terms interchangably.

## 5 Evaluation results

In this section, we first present the evaluation results to show that our curriculum adversarial training approach can improve the resilience over [Madry *et al.*, 2018] on different tasks and different models. Then, we will perform ablation studies to study the effectiveness of each component. We will also document the forgetting phenomenon and the attack generalization issue which motivate the design of quantization and batch mixing respectively.

### 5.1 Resilience comparison

In this subsection, we first compare our curriculum adversarial training strategy against [Madry *et al.*, 2018]. The results are presented in Table 1. We can observe that on all datasets, our CAT approah can outperform the prior-art. Especially, on CIFAR-10 and SVHN, the margin ranges from 25% to 35%. Note that on CIFAR-10, the best model reported in [Madry *et al.*, 2018] is WideResNet, which can achieve 47% empirical worst-case accuracy. Comparing with this result, applying our CAT technique to DenseNet-161 can still improve it by 24%.

On SVHN, we observe that ResNet-50 trained with [Madry *et al.*, 2018] is unexpectedly bad at defending against adversarial examples. We observe that its training accuracy is not improving over a long period of time. We conduct sanity checks of our implementation, and find no obvious bugs. One potential reason is that ResNet-50 may not be sufficiently large.

We notice that our replication of [Madry *et al.*, 2018] using DenseNet-161 can achieve an accuracy as high as 88% under attack PGD(10), which is used during adversarial training. However, we observe that the model achieving 88% under PGD(10) performs poorly on stronger attacks: the accuracy drops to 9% for PGD(50) and 5% for PGD(100). As a result, the empirical worst-case accuracy of this model, which overfits to PGD(10), is lower than 5%. In light of this issue, we obtain the model achieving the best empirical worst-case accuracy on the validation set during training, rather than the model overfitting to PGD(10). In doing so, we observe that this model's accuracy under all strong attacks (i.e., PGD($k$) with $k \geq 3$) is 40% to 50%; and as a result, the empirical worst-case accuracy is 40.38% as reported in Table 1. With respect to this result, our evaluation shows that our CAT approach can improve [Madry *et al.*, 2018] by a margin of 35%.

The only other independent replication of [Madry *et al.*, 2018] is from [Jacob Buckman, 2018], who report an accuracy of [Madry *et al.*, 2018] under PGD(10) to be 59.63%. One may wonder whether [Jacob Buckman, 2018] achieve better resilience on SVHN. While they claim they achieve a better performance under white-box attacks (i.e., 94.77%), [Jacob Buckman, 2018]'s accuracy against black-box attacks is very low (i.e., 48.67%). Thus, based on our metric, their approach's empirical worst-case accuracy is at most 48.67%; and our approach outperforms [Jacob Buckman, 2018] by at least, i.e., 27%.

It is worth to mention that, on SVHN, a concurrent work, [Kolter and Wong, 2017], provides a model with provable resilience of at least 59.33%. The empirical worst case accuracy of their approach is at most 65.48%; thus our approach outperforms it by a margin of at least 10%.

### 5.2 Detailed analysis

In this section, we present some detailed analysis of the proposed techniques over previous approach. Most of the discussion will be based on ablation study results presented in Table 2. In the following, we will discuss the results with respect to various questions.

**The effectiveness of CAT.** We first document the effectiveness of basic curriculum adversarial training to overcome the overfitting problem of [Madry *et al.*, 2018]. In particular, we show the validation and test empirical worst-case accuracy over time of both our approach and [Madry *et al.*, 2018].

In Figure 1, we plot the empirical worst-case accuracy on the test data (i.e., test resilience) of both our CAT ap-

### CIFAR-10

|  | AT | AT+Quant | Basic | Basic+Quant | MIX | MIX+Quant | Basic+MIX | CAT |
| --- | --- | --- | --- | --- | --- | --- | --- | --- |
| ResNet-50 | 45.39% | 46.27% | 41.09% | 62.78% | 38.69% | 39.90% | 2.85% | 65.93% |
| DenseNet-161 | 46.18% | 47.04% | 41.40% | 61.25% | 33.82% | 35.33% | 1.20% | 69.27% |

### SVHN

|  | AT | AT+Quant | Basic | Basic+Quant | MIX | MIX+Quant | Basic+MIX | CAT |
| --- | --- | --- | --- | --- | --- | --- | --- | --- |
| ResNet-50 | 19.59% | 19.59% | 0.00% | 62.39% | 32.65% | 72.67% | 0.43% | 75.66% |
| DenseNet-161 | 40.38% | 61.86% | 0.01% | 67.79% | 29.03% | 41.88% | 0.25% | 74.58% |

Table 2: Ablation study. Empirical worst-case accuracy is reported. "AT" indicates vanilla adversarial training approach. "Basic" indicates using the curriculum; "Quant" indicates using quantization; and "MIX" indicates using batch mixing.

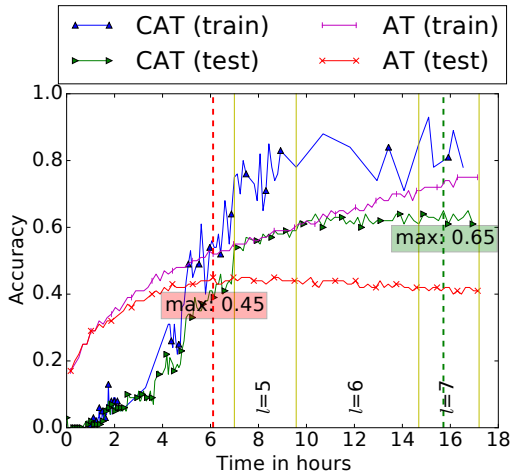

Figure 1: Training and testing empirical worst-case accuracy of vanilla adversarial training and curriculum adversarial training over time. The model is ResNet-50, and the dataset is CIFAR-10.

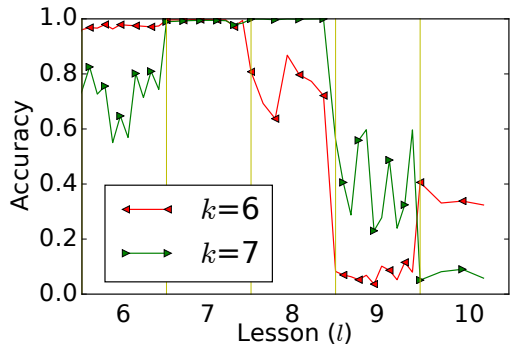

Figure 2: Catastrophic forgetting phenomenon on SVHN using ResNet-50. The plots show the accuracy under PGD attacks under different strength. We use basic curriculum adversarial training to train the model. The x-axis correspond to the different lessons in the curriculum.

proach and [Madry *et al.*, 2018]'s approach. The validation resilience's figure is similar, and thus we omit it. From the figure, we can observe that AT soon achieves a better test resilience, because from the very beginning, the model sees adversarial examples generated from strong attacks; but it hits the peak at around $45\%$ and then starts dropping. At the same time, we observe that the training resilience curve of AT continues increasing. Therefore, *AT overfits to the training data, and does not generalize to the test data*.

On the other hand, we observe that CAT's test resilience bypasses AT about the same time as AT hits the peak after simpler lessons (i.e., $l = 4$) are finished, and continues to increase to above $60\%$. Therefore, we conclude that our CAT technique can effectively overcome the overfitting problem of AT.

**The forgetting phenomenon.** During basic curriculum adversarial training, we observe that when the model finishes lesson $l$, the model achieves the highest accuracy under attack $k = l$. However, when continuing to train with harder lessons, the accuracy under attack PGD($k$) may drop significantly. Figure 2 illustrates such an example on ResNet-50 models on SVHN dataset. We can observe that the accuracy against PGD(6) and PGD(7) increases to almost $100\%$ after lesson $l = 7$; but when training with lesson $l = 9$, the accuracy of attack PGD(6) drops to below $10\%$. This illustrate that the model entirely forgets its training on lesson $l = 6$, and thus we refer to this as a *catastrophic forgetting phenomenon*. The same phenomenon can be observed with PGD(7) attack as well.

**The effectiveness of batch mixing.** We now show that batch mixing can effectively mitigate the issue caused by forgetting. We present the $\tilde{l}$-accuracy defined by (5) using basic curriculum adversarial training (Basic) and basic curriculum adversarial training enhanced with batch mixing (Basic+MIX) in Table 3 for comparison. We can observe that Basic+MIX can consistently improve the performance over Basic by a margin ranging from $5.88\%$ to $68.3\%$. Especially, on SVHN, the reason why the empirical worst-case accuracy of Basic is as low as around $10\%$ is because of the forgetting phenomenon as documented above. Therefore, our batch mixing technique can effectively mitigate this issue. Also, we can observe from Table 3 that Basic+MIX can outperform AT on $\tilde{l}$-accuracy as well. These results demonstrate the effectiveness of batch mixing.

However, if we examine the empirical worst-case accuracy in Table 2, we observe that the performance of Basic+MIX is extremely bad, i.e., less than $6\%$. This is mainly because of the attack generalization issue that will be discussed next.

|        | CIFAR-10 |          | SVHN   |          |
|--------|----------|----------|--------|----------|
|        | ResNet   | DenseNet | ResNet | DenseNet |
| AT     | 45.78%   | 46.67%   | 19.59% | 55.98%   |
| Basic  | 67.48%   | 63.85%   | 10.72% | 8.25%    |
| Basic+MIX | 73.36% | 73.05%  | 79.02% | 65.16%   |

Table 3: $\tilde{l}$-accuracy for $l = 7$ (for CIFAR-10) and $l = 10$ (for SVHN)

|         | CIFAR-10 |          | SVHN   |          |
|---------|----------|----------|--------|----------|
|         | ResNet   | DenseNet | ResNet | DenseNet |
| Regular | 94.31%   | 95.70%   | 95.66% | 96.65%   |
| CAT     | 88.49%   | 90.36%   | 95.30% | 95.20%   |

Table 4: Evaluation results on pristine test set. ResNet indicates ResNet-50; DenseNet indicates DenseNet-161. All results (including CAT) are the test accuracy on non-adversarial inputs.

By including quantization (+Quant) to mitigate this issue, we again observe that Basic+MIX+Quant, which is equivalent to the full CAT approach can outperform all other baselines.

**The attack generalization issue.** As we have discussed above, by using curriculum adversarial training with batch mixing, the $\tilde{K}$-accuracy is high, which indicates that the model can effectively defend against all attacks weaker than PGD($K$). However, we observe in various places that for attacks stronger than PGD($K$), the accuracy may drop significantly. Therefore, *a model trained with weaker attacks may not generalize to stronger attacks*. We refer to this effect as *the attack generalization issue*.

One naive idea to mitigate this issue is to increase the upper bound on $K$; however, this idea clearly cannot solve the problem since we cannot train with an infinitely large $K$. Therefore, no matter how large we choose $K$, there may always be stronger attacks than PGD($K$) that can break the defense.

**The effectiveness of quantization.** Since quantization is an inference time defense, it can be used by any combinations of models, datasets, and training techniques. In Table 2, we observe that quantization (+Quant) improves the performance on almost all such combinations. The reason is because quantization can effectively reduce the attack space, so that even stronger attacks cannot find more sub-optimal solutions to the attack optimization problem (2) than weaker attacks when the input pixel values are quantized to a few bits. From Table 2, we also observe that quantization also helps to improve the resilience of AT from [Madry *et al.*, 2018]. Therefore, we confirm that quantization is a generic defense tool to work with a wide range of other defense techniques.

**Does batch mixing replace the role of curriculum?** From above, we have observed that batch mixing can help curriculum adversarial training. A natural question is that whether the curriculum is helpful or not? In fact, since it makes the model be supervised with both adversarial examples generated from weaker attacks and from stronger ones, batch mixing may potentially replace the role of the curriculum. To answer this, in Table 2, we report this comparison by showing the resilience by using the curriculum (i.e., Basic) or not. The most significant phenomenon is that CAT, which is equivalent to Basic+MIX+Quant, outperforms batch mixing with quantization but without curriculum (MIX+Quant) by a large margin in most cases. This clearly demonstrates that the curriculum is important to achieve a high resilience.

However, we also observe that compared with MIX only, Basic+MIX's performance is much worse. The reason has been explained above: the model trained with Basic+MIX is overfitting to PGD($k$) attacks for $k \leq K$, but is not resilient at all to stronger attacks. In comparison, the model trained with MIX sacrifices some resilience against with weaker attacks to trade for more resilience against stronger attacks. When quantization is deployed to effectively mitigate stronger attacks, and thus the trade-off made by MIX is less necessary. This is why we can observe a better overall resilience of CAT. In summary, we conclude that the curriculum is a useful procedure to achieve high resilience, but it is better to be used with batch mixing and quantization together.

**Does CAT hurt regular accuracy?** The last question is whether adversarial training hurts the accuracy on non-attack inputs. We conduct experiments to compare our CAT's accuracy with the same model trained using their standard routine on the same dataset. The results are presented in Table 4.

For SVHN, the accuracy drops around $1\%$; for CIFAR-10, the drops are higher, but still within the range of $5\%$ to $6\%$. We attribute this to the fact that our training curriculum and batch mixing consider non-adversarial inputs during training. Such a performance loss is acceptable in many practical scenarios in trade of a better resilience.

# 6 Conclusion

In this work, we present curriculum adversarial training as a defense mechanism to defend against adversarial example attacks. It follows the adversarial training framework, which has recently been shown effective by [Madry *et al.*, 2018]. We make several innovations to significantly improve the algorithm's resilience. In particular, we propose a technique of adversarial training to follow a curriculum containing adversarial examples generated by attacks with various strength; and we also show two effective optimizations called batch mixing and quantization. By combining all the techniques, we demonstrate that we can improve over previous state-of-the-art by a margin ranging from $25\%$ to $35\%$. Our work sheds new light on further investigation directions on adversarial training frameworks.

## Acknowledgement

This material is in part based upon work supported by DARPA award under no. FA8750-17-2-0091, Center for Long-Term Cybersecurity, and Berkeley Deep Drive.